\documentclass{article}
\IfFileExists{PRIMEarxiv.sty}{\usepackage{PRIMEarxiv}}{\usepackage[letterpaper,margin=1in]{geometry}}
\usepackage[utf8]{inputenc}
\usepackage[authoryear,round]{natbib}
\usepackage[T1]{fontenc}
\usepackage{hyperref}
\usepackage{url}
\usepackage{booktabs}
\usepackage{amsfonts,amsmath,amssymb}
\usepackage{nicefrac}
\usepackage{microtype}
\usepackage{fancyhdr}
\usepackage{graphicx}
\usepackage{array}
\usepackage{adjustbox}
\usepackage{algorithm}
\usepackage{algpseudocode}
\graphicspath{{media/}}
\title{Atomic MathReasoner: Instance-Level Composition of Atomic Reasoning Skills for Mathematical LLMs}

\author{
Jiashu Li, Tianyu Gao, Qinglong Liu, Jinping Wang, Ye Wei, Zhiqiang Gao
}

\date{}
\begin{document}
\maketitle
\begin{abstract}

Most existing approaches enhance mathematical reasoning in large language models (LLMs) as a monolithic capability, despite the fact that different mathematical problems require different reasoning skills. Recent work has explored a more fine-grained perspective by decomposing mathematical reasoning into atomic skills and developing specialized modules for individual capabilities. However, how to effectively utilize these atomic abilities remains underexplored. To address this challenge, we propose Atomic MathReasoner (AMR), an instance-level framework for composing atomic reasoning skills. AMR first trains a library of capability-specific LoRA experts and performs gradient-free optimization over held-out validation subsets to discover a diverse set of low-loss composition prototypes. It evaluates the preference of each prototype using the token-level negative log-likelihood of reference solutions, filters ambiguous assignments through a confidence margin, and distills the resulting pseudo-labels into a lightweight router. During inference, the router selects an appropriate composition prototype for each problem, enabling problem-specific utilization of atomic skills without costly online optimization. Experiments on the AtomicMath benchmark across general and mathematics-specialized LLMs demonstrate consistent improvements. On Qwen2.5-Math-Instruct-7B, AMR increases average field accuracy from 64.4\% to 68.1\% at Level 1 and from 59.3\% to 63.7\% at Level 2, while improving average logical reasoning performance from 39.4\% to 46.8\%. These results demonstrate that effectively composing specialized atomic skills provides a promising approach to strengthening mathematical reasoning in LLMs.

\end{abstract}
\section{Introduction}

Mathematical reasoning is a fundamental capability of artificial intelligence \citep{azerbayev2024llemma}. Large language models (LLMs) have achieved substantial progress in mathematical problem solving. Recent advances have been driven by several paradigms, including instruction tuning \citep{lewkowycz2022solving,luo2023wizardmath,yu2024metamath,yue2024mammoth}, tool-integrated reasoning \citep{chen2022program,gao2023pal,gou2024tora,chen2026toward}, and reinforcement learning over long chains of thought \citep{lai2024step,wang2024math,liao2025rewardguided,guo2025deepseek,yan2026mathfimer}. Despite their effectiveness, most of these paradigms treat mathematical reasoning as a monolithic capability, optimizing either the entire model or the complete solution trajectory.

This limitation motivates a more fine-grained view of mathematical reasoning. Human problem solving often decomposes complex tasks into simpler sub-skills and coordinates them according to problem-specific demands \citep{Anderson01041995,Zhang1999SourcesOM}. Inspired by this perspective, \citet{kuang2026atomic} decomposed mathematical reasoning in LLMs into fine-grained atomic skills, including domain-specific and logical reasoning capabilities.

However, decomposing mathematical reasoning into atomic skills does not by itself determine how these skills should be combined for a specific problem. A central challenge therefore remains at inference time: how can a model select and compose the appropriate atomic skills for an unseen input? Although \citet{kuang2026atomic} showed that activating relevant atomic skills can improve performance, their work primarily focused on capability decomposition and did not provide an efficient mechanism for skill composition.

To bridge this gap, we introduce Atomic MathReasoner (AMR), a two-stage framework for instance-adaptive composition of atomic mathematical skills. In the first stage, following the atomic-skill taxonomy of \citet{kuang2026atomic}, we train a dedicated LoRA expert for each skill. We then conduct multiple gradient-free searches on a held-out validation set to optimize the linear composition coefficients under the causal language modeling loss, yielding a diverse set of compositional prototypes. In the second stage, we formulate problem-specific prototype selection as a classification task, avoiding costly per-instance search during inference. For each routing example, we assign the prototype with the lowest token-level negative log-likelihood on the reference solution as a pseudo-label and apply margin-based filtering to retain high-confidence examples. We subsequently train a lightweight router based on TF-IDF features and multinomial logistic regression. At inference time, the router maps each unseen problem to a compositional prototype, whose coefficients are used to combine the atomic LoRA experts for answer generation.

Our extensive experiments on the AtomicMath benchmark \citep{kuang2026atomic} demonstrate the robust effectiveness of AMR across various base models. Specifically, AMR improves the average Level 1 and Level 2 field accuracies of InternLM2-Math-Plus-7B to 42.1\% (+4.3 percentage points) and 38.7\% (+2.2 percentage points), respectively, while increasing the average logic score of Qwen2.5-Math-Instruct-7B by 7.4 percentage points to 46.8\%. Notably, our framework enables smaller models to outperform larger math-specialized counterparts in specific categories. For instance, InternLM2-Math-Plus-7B with AMR achieves 48.4\% in Analysis (Level 1) and 38.8\% in Topology (Level 2), surpassing Abel-13B (40.4\%) and Xwin-Math-13B (38.3\%), respectively. AMR also remains competitive with commercial models on several challenging tasks. In particular, Qwen2.5-Math-Instruct-7B with AMR achieves 73.5\% in Analysis (Level 2) and 53.8\% in Geometry (Level 2), outperforming GPT-4o (53.2\% and 36.3\%) and Claude 3.7 Sonnet (68.1\% and 39.0\%) on these two tasks, respectively. Additionally, a comprehensive ablation study further validates the effectiveness of the proposed framework.

\begin{itemize}

\item We propose AMR, a two-stage framework that composes independently trained atomic skill experts at the instance level. By constructing problem-conditioned skill compositions, AMR enables different mathematical problems to leverage complementary reasoning capabilities.

\item We propose a gradient-free composition search method and lightweight offline router. By distilling prototype labels into the routing classifier, AMR enables efficient instance-level capability matching during inference and reduces latency.

\item Our framework, when integrated into diverse baseline models, demonstrates performance gains across both low and high difficulty levels on the AtomicMath benchmark. It enables 7B models to exhibit strong competitiveness against 13B specialized counterparts and even GPT-4o in specific domains.

\end{itemize}

\begin{figure}[t]
 \centering
 \includegraphics[width=0.96\linewidth]{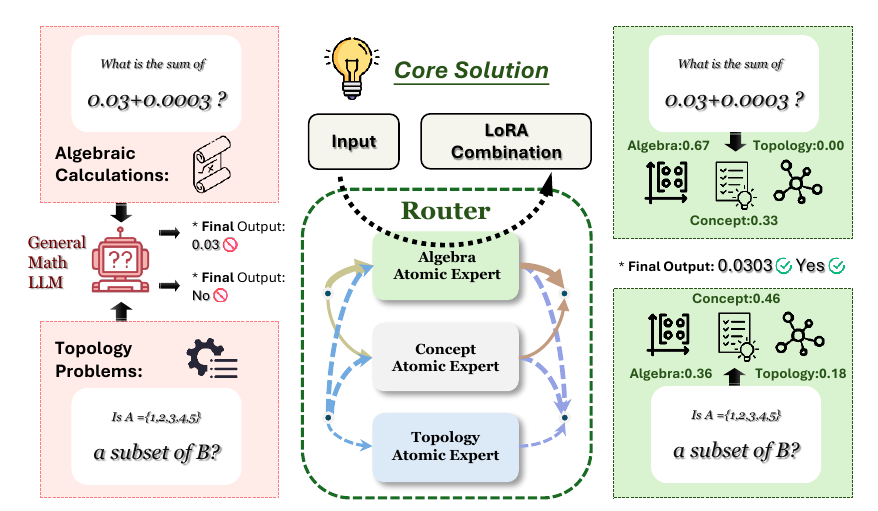}
 \caption{Conventional LLMs treat mathematical reasoning as a monolithic capability, often failing across diverse mathematical problems. In contrast, AMR dynamically composes atomic reasoning experts for each input through instance-level routing, enabling problem-adaptive capability synergy and more accurate mathematical reasoning.}
 \label{fig:motivation}
\end{figure}

\section{Related Work}

\subsection{Mathematical Reasoning in LLM}

Mathematical reasoning for Large language models (LLMs) has achieved substantial progress through several complementary paradigms including chain-of-thought and self-consistency decoding \citep{cot,cot2}, domain-specific pretraining and instruction tuning on mathematical corpora \citep{lewkowycz2022solving,luo2023wizardmath}, tool-integrated reasoning that delegates computation to external programs or symbolic solvers \citep{gao2023pal,chen2022program,gou2024tora}, and reinforcement-learning-based reasoning models \citep{lai2024step,wang2024math,liao2025rewardguided,guo2025deepseek,yan2026mathfimer}. Recent work on Atomic Thinking argues that mathematical intelligence can be decomposed into fine-grained atomic capabilities \citep{kuang2026atomic}. However, decoupling abilities alone does not determine how these heterogeneous skills should be composed for each unseen problem. Our work addresses this gap by dynamically composing atomic mathematical LoRA experts at the instance level.

\subsection{Skill Decomposition and Expert Composition}

Decomposing complex reasoning tasks into simpler subproblems or specialized skills has been shown to improve performance on challenging tasks \citep{xiang2025atomic,teng2025atom,shi2022skill,jiang2026scribe,zhang2026atomicvla}. This perspective is also consistent with cognitive accounts in which complex problem solving involves coordinating multiple specialized capabilities \citep{Anderson01041995,Zhang1999SourcesOM}. While these studies primarily investigate task decomposition or the construction of intermediate subgoals, AMR focuses on the complementary problem of composing a fixed library of independently trained atomic mathematical experts for each input. From a methodological perspective, AMR is related to parameter-efficient adapter composition \citep{pfeiffer2021adapterfusion}, mixture-of-experts routing \citep{fedus2022switch}, and model merging \citep{wang2026escaping}, which investigate how specialized modules or parameter updates can be combined or selected. AMR applies this general composition perspective to atomic mathematical skills and studies whether problem-conditioned expert mixtures can improve mathematical reasoning.

\section{Methodology}

In this section, we introduce the three main components of AMR. We first construct a library of atomic LoRA experts based on fine-grained mathematical capabilities (\textbf{LoRA for Each Atomic Ability}). Next, we build a compositional prototype set through gradient-free search on the validation set (\textbf{Composition Prototype Set Construction}). Finally, we generate prototype-based pseudo-labels and train a lightweight router to select a suitable expert composition for each input during inference (\textbf{Instance-Level Routing via Prototype-Based Pseudo-Labels}).

\begin{figure}[t]
\centering
\includegraphics[width=0.98\linewidth]{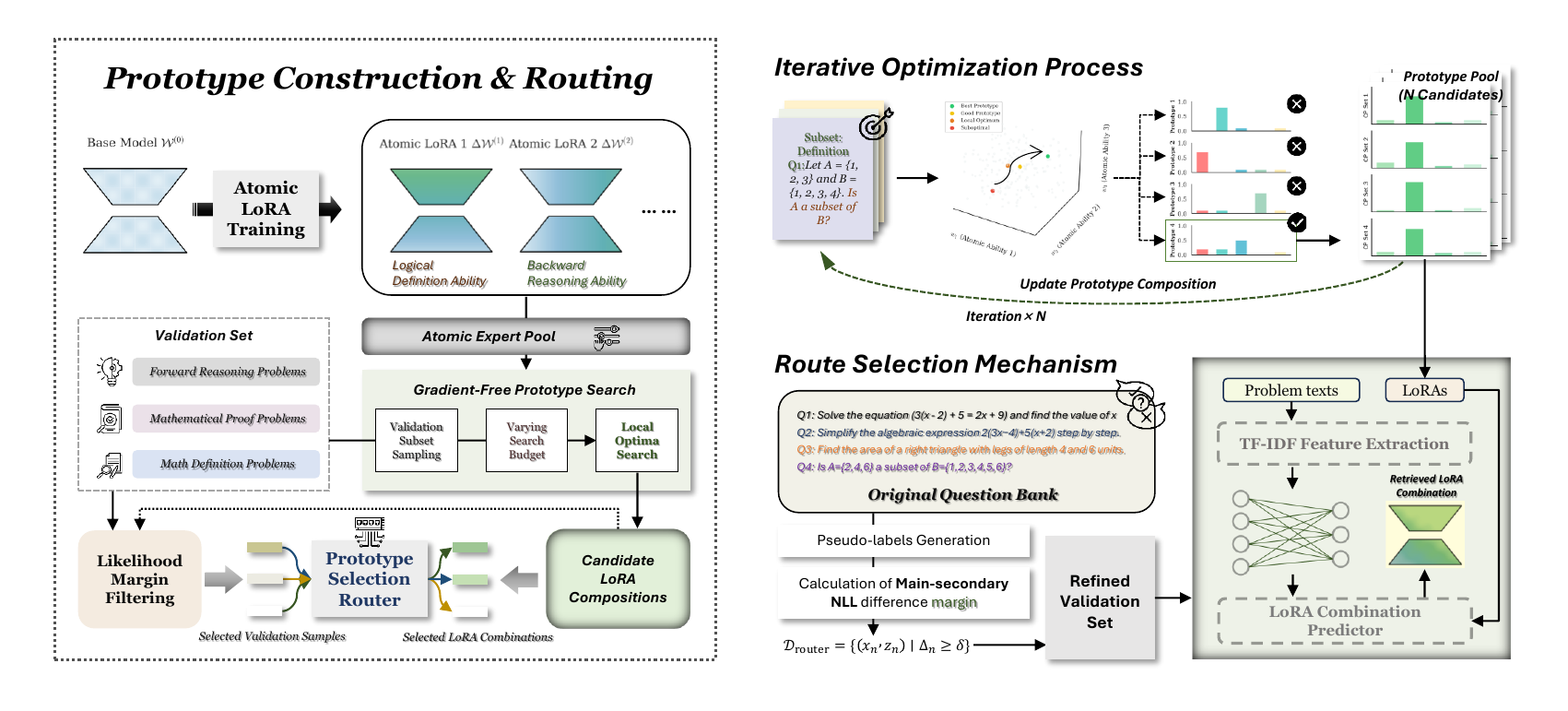}

\caption{Overview of the proposed Atomic MathReasoner (AMR) framework. Top (Stage 1): Prototype Construction. We first decouple mathematical reasoning into atomic capabilities and train corresponding LoRA modules. A gradient-free optimization algorithm is then applied to the validation set to search for localized capability compositions, constructing a Compositional Prototype Set. Bottom (Stage 2): Route Selection Mechanism. We generate pseudo-labels via token-level NLL and apply a margin-based filtering strategy to obtain a noise-removed training set. Finally, a lightweight router utilizing TF-IDF feature extraction is trained to dynamically predict a suitable combination for unseen problems.}

\label{fig:framework}
\end{figure}

\subsection{LoRA for Each Atomic Ability}

Following the taxonomy of mathematical reasoning abilities defined by \citet{kuang2026atomic}, we denote the set of $N$ atomic abilities as $\mathcal A=\{a_1,a_2,\ldots,a_N\}$. To ensure a clear decoupling of these distinct cognitive skills, we train a dedicated LoRA module $\psi_i$ \citep{hu2022lora} for each corresponding atomic ability $a_i\in\mathcal A$. Specifically, each module is fine-tuned independently using the specialized datasets provided by \citet{kuang2026atomic}. This process results in a comprehensive library of LoRAs, where each $\psi_i$ encapsulates the specific atomic ability of its corresponding task.

\subsection{Composition Prototype Set Construction}

Each atomic LoRA module has its specific mathematical knowledge. However, solving real mathematical problems requires the dynamic synergy of these capabilities. For instance, an analytic geometry problem demands a deep integration of both algebraic manipulation and spatial geometric intuition, whereas a pure algebraic equation requires concentrated algebraic reasoning. A static combination such as applying fixed global weights to all problems fails to adapt to the diverse demands of different questions, often introducing irrelevant noise.

To overcome the limitation that a single composition strategy struggles to adapt to diverse mathematical problems, we discard the reliance on a unique global compositional solution. Instead, we collect a diverse set of compositional prototypes through multiple gradient-free searches. Specifically, given a designated search subset of the validation set, $S_{\mathrm{search}}$, we search for weight vectors $\mathbf w=[w_1,\ldots,w_N]^\top$ within a continuous bounded search space $\Omega\subset\mathbb R^N$. For a pre-trained weight $W^{(0)}$ of a target linear layer in the base model, we denote the low-rank update corresponding to the $i$-th atomic LoRA module as $\Delta W^{(i)}=B^{(i)}A^{(i)}$, and adopt a linear incremental composition:

\begin{equation}
W(\mathbf{w})=W^{(0)}+\sum_{i=1}^{N} w_i\Delta W^{(i)}
=W^{(0)}+\sum_{i=1}^{N}w_i B^{(i)}A^{(i)}.
\label{eq:composition}
\end{equation}

Subsequently, we employ a gradient-free optimization algorithm (e.g., NGOpt) to search for the composition coefficients that minimize the validation loss:

\begin{equation}
\mathbf w^*=\arg\min_{\mathbf w\in\Omega}
\left\{\mathcal{L}_{\mathrm{val}}(\mathbf w)+\lambda\frac{1}{N}\sum_{i=1}^N|w_i|\right\}.
\label{eq:search}
\end{equation}

where $\mathcal L_{\mathrm{val}}(\mathbf w)$ denotes the causal language modeling loss evaluated on the validation samples, and $\lambda$ is the L1 sparsity regularization coefficient. By repeatedly executing this process across different support subsets of the validation set and under varying search budget settings, we obtain multiple validation-loss local optima in the composition space. After deduplicating the repeated solutions, these vectors constitute the compositional prototype set:

\begin{equation}
\Pi=\{\pi_1,\ldots,\pi_M\}.
\label{eq:prototype}
\end{equation}

where each prototype $\pi_m$ corresponds to an atomic capability synergy pattern.

\subsection{Instance-Level Routing via Prototype-Based Pseudo-Labels}

While the prototype set constructed in the previous stage encapsulates diverse synergistic strategies, dynamically matching an unseen problem to a suitable composition remains a challenge. To establish a fast and direct mapping from the problem text to a composition, we frame instance-level prototype selection as a classification task.

Given the compositional prototype set $\Pi=\{\pi_1,\ldots,\pi_M\}$, where each prototype corresponds to a fixed vector of composition weights, we generate prototype-based pseudo-labels by evaluating the token-level negative log-likelihood (NLL) for each input-output pair $(x_n,y_n)$ in a routing subset of the held-out validation set $S_{\mathrm{route}}$, which is disjoint from the prototype-search subset $S_{\mathrm{search}}$. The prototype achieving the lowest NLL is assigned as the prototype-based pseudo-label $z_n$:

\begin{equation}
z_n=\arg\min_{m\in\{1,\ldots,M\}}\mathcal{L}_{\mathrm{NLL}}(x_n,y_n;\pi_m).
\label{eq:pseudolabel}
\end{equation}

We use this NLL-selected label as a tractable proxy for prototype preference, rather than assuming that it is the unique semantic optimum for the problem. Directly learning from these pseudo-labels introduces noise, as multiple prototypes may yield comparable NLLs for a given problem. To address this, we introduce a margin-based filtering mechanism that governs a critical trade-off between label purity and sample coverage.

Let $\ell_n^{(1)}$ and $\ell_n^{(2)}$ denote the lowest and second-lowest NLLs across all prototypes. We define the margin $\Delta_n=\ell_n^{(2)}-\ell_n^{(1)}$. By enforcing a margin threshold $\delta>0$, we filter out ambiguous samples and construct a refined training set:

\begin{equation}
\mathcal D_{\mathrm{router}}=\{(x_n,z_n)\mid\Delta_n\ge\delta\},
\qquad\Delta_n=\ell_n^{(2)}-\ell_n^{(1)}.
\label{eq:routerdata}
\end{equation}

A properly tuned $\delta$ ensures the router learns from high-confidence mappings without excessively truncating the problem-type distribution. To avoid introducing excessive latency during testing, we do not employ neural network routers such as Transformer encoders. Instead, we extract non-parametric TF-IDF features $\phi(x_n)$ from the text and train a lightweight multinomial logistic regression classifier.

The classifier is optimized using standard cross-entropy loss with $L_2$ regularization over $\mathcal D_{\mathrm{router}}$. During inference, for any unseen problem $x$, the router predicts a prototype index in negligible time. The corresponding weight vector is then retrieved to linearly combine the atomic LoRA modules. Ultimately, this paradigm replaces costly online dynamic searches with a highly efficient offline-trained classifier, allowing each problem to use a composition of reasoning abilities.

\begin{algorithm}[t]
\caption{Inference Pipeline of AMR}
\label{alg:inference}
\begin{algorithmic}[1]
\Require Problem $x$, base weights $W^{(0)}$, atomic LoRAs $\{\Delta W^{(i)}\}_{i=1}^{N}$, prototypes $\Pi$, trained router
\State $\phi(x)\gets \mathrm{TFIDF}(x)$
\State $m^*\gets \mathrm{Router}(\phi(x))$
\State $\mathbf{w}^*\gets\pi_{m^*}$
\State $W(\mathbf{w}^*)\gets W^{(0)}+\sum_i w_i^*\Delta W^{(i)}$
\State $y\gets\mathrm{LLM.Generate}(x\mid W(\mathbf{w}^*))$
\State \Return $y$
\end{algorithmic}
\end{algorithm}

\section{Experiment}

\subsection{Experimental Details}

\paragraph{Baseline.} In this paper, we employ Llama-3-8B-Instruct \citep{llama3modelcard} as a general LLM, and InternLM2-Math-Plus-7B \citep{ying2024internlmmath} and Qwen2.5-Math-Instruct-7B \citep{yang2024qwen2} as math-specialized LLMs, to serve as the baselines for applying our method. Furthermore, we select diverse open-source and state-of-the-art closed-source LLMs for comparison; detailed in Appendix. For reasoning enhancement frameworks, we compare AMR with methods which aim to improve mathematical reasoning, including DPO \citep{lai2024step}, Step-DPO \citep{lai2024step}, Math-Shepherd \citep{wang2024math}, RSD \citep{liao2025rewardguided}, MathFimer \citep{yan2026mathfimer}. Datasets. We evaluate performance on AtomicMath benchmark \citep{kuang2026atomic}, which includes both field and logical dimensions. For field capabilities, we collect and categorize data from MATH \citep{hendrycksmath2021}, GSM8K \citep{cobbe2021gsm8k}, Gaokao-Bench \citep{Zhang2023EvaluatingTP}, OlympiadBench \citep{he2024olympiadbench}, AIME, MMLU \citep{wang2024mmlu}, and DeepMath \citep{he2026deepmathk} into four fields (algebra, geometry, analysis, topology) across two difficulty levels. For logical capabilities, we derive conceptual understanding tasks from NaturalProofs \citep{welleck2021naturalproofs}, formal language forward reasoning tasks from Lean Workbook, and counterexample-driven backward reasoning statements from CounterMath \citep{li2025one}. Notably, we select GSM8K \citep{cobbe2021gsm8k} and MATH \citep{hendrycksmath2021} for comparison with other methods for mathematical reasoning enhancement. We follow the official AtomicMath train/test split and we hold out 10\% training data as validation. The training and validation splits are disjoint from the test split: atomic LoRA experts are trained only on the training split, prototype search and router fitting are performed only on the validation split, and the test split is reserved strictly for evaluation. Evaluation and Metrics. Following \citet{kuang2026atomic}, we adopt default Chain-of-Thought (CoT) prompting for all evaluations. For field capabilities, conceptual understanding, and forward reasoning tasks, we use Accuracy computed via exact match with reference answers. For backward reasoning, we evaluate truthfulness judgment using the F-1 score. Additionally, to assess counterexample construction, we use Example extraction alongside Strict and Loose alignment metrics to measure consistency with reference examples.

\paragraph{Implementation Details.} We train a separate atomic LoRA adapter for each capability using LLaMA-Factory \citep{zheng2024llamafactory}. To ensure a fair comparison, all atomic adapters share the same training recipe, including 4-bit quantization, LoRA rank 16, batch size of 16, learning rate $10^{-4}$, and 2 training epochs. For the gradient-free composition search, the continuous bounded search space $\Omega$ is set to $[-1.5,1.5]^N$ (where N = 7 represents the number of decoupled atomic experts), and the L1 sparsity regularization coefficient $\lambda$ is set to 0.05. Unless otherwise specified, the router uses all M = 9 prototypes and a margin threshold of $\delta$ = 0.02. We sample 512 examples from validation split as the candidate pool for pseudo-label assignment; after margin filtering, the retained high-confidence examples are used to train the router. All composition search, prototype extraction, and router fitting are conducted exclusively on the validation split, while the test split is used only for final evaluation. Unless otherwise specified, all reported results are obtained from three runs under three random seeds. More details are in Appendix.

\begin{table}[htbp]\centering\caption{Performance of AMR on the field atomic capabilities of AtomicMath (three runs). Values are accuracies under low and high difficulty.}\label{tab:field}

\begin{adjustbox}{max width=\textwidth}\begin{tabular}{l*{10}{r}}\toprule

Method & \multicolumn{2}{c}{Algebra} & \multicolumn{2}{c}{Geometry} & \multicolumn{2}{c}{Analysis} & \multicolumn{2}{c}{Topology} & \multicolumn{2}{c}{Avg.} \\

 & Low & High & Low & High & Low & High & Low & High & Low & High \\\midrule

Llama-3-8B-Instruct & 48.2 & 28.7 & 30.5 & 24.3 & 39.1 & 27.5 & 22.0 & 20.1 & 35.0 & 25.2 \\

InternLM2-Math-Plus-7B & 49.2 & 35.9 & 33.0 & 31.4 & 41.9 & 41.5 & 27.2 & 37.0 & 37.8 & 36.5 \\

Deepseek-math-RL-7B & 52.0 & 33.8 & 35.5 & 39.3 & 44.6 & 37.6 & 22.6 & 33.7 & 38.7 & 36.1 \\

NuminaMath-7B-TIR & 52.4 & 28.2 & 39.9 & 32.1 & 46.8 & 33.6 & 26.4 & 25.1 & 41.4 & 29.8 \\

Qwen2.5-Math-Instruct-7B & 80.5 & 65.2 & 57.3 & 51.9 & 67.7 & 66.5 & 52.1 & 53.4 & 64.4 & 59.3 \\

Abel-13B & 63.4 & 47.3 & 42.7 & 41.8 & 40.4 & 35.6 & 30.2 & 24.6 & 44.2 & 37.3 \\

Xwin-Math-13B & 78.3 & 51.6 & 49.5 & 46.7 & 55.7 & 54.8 & 46.7 & 38.3 & 57.6 & 47.9 \\

InternLM2-Math-Plus-20B & 67.8 & 49.6 & 49.3 & 46.1 & 54.3 & 47.8 & 45.4 & 38.9 & 54.2 & 45.6 \\

GPT-4o & 69.3 & 50.5 & 48.5 & 36.3 & 55.1 & 53.2 & 53.0 & 49.4 & 56.5 & 47.4 \\

Claude 3.7-Sonnet & 83.5 & 72.0 & 58.0 & 39.0 & 80.3 & 68.1 & 56.0 & 58.6 & 69.5 & 59.4 \\

Gemini 2.5-Pro & 80.5 & 79.6 & 70.2 & 73.1 & 67.5 & 63.8 & 68.1 & 66.2 & 71.6 & 70.7 \\

\midrule

Llama-3 + AMR (Ours) & 54.3 & 33.2 & 31.7 & 25.4 & 45.2 & 31.5 & 23.1 & 21.4 & 38.6 & 27.9 \\

InternLM2 + AMR (Ours) & 55.1 & 40.3 & 35.4 & 33.1 & 48.4 & 42.6 & 29.5 & 38.8 & 42.1 & 38.7 \\

Qwen2.5 + AMR (Ours) & 84.1 & 71.4 & 59.9 & 53.8 & 74.3 & 73.5 & 53.9 & 55.9 & 68.1 & 63.7 \\

\bottomrule\end{tabular}\end{adjustbox}\end{table}

\begin{table}[htbp]\centering\caption{Logic atomic capabilities of AtomicMath (three runs); accuracy for concept understanding and forward reasoning, F1 for backward reasoning.}\label{tab:logic}

\begin{adjustbox}{max width=\textwidth}\begin{tabular}{lrrrrr}\toprule

Method & Concept Attr. & Concept Def. & Forward Acc. & Backward F1 & Avg. \\\midrule

Llama-3-8B-Instruct & 42.1 & 35.2 & 20.5 & 24.3 & 30.5 \\

InternLM2-Math-Plus-7B & 46.2 & 43.2 & 23.7 & 33.9 & 36.8 \\

Deepseek-math-RL-7B & 46.5 & 39.4 & 23.1 & 27.6 & 34.2 \\

NuminaMath-7B-TIR & 27.3 & 22.8 & 30.2 & 30.4 & 27.7 \\

Qwen2.5-Math-Instruct-7B & 50.3 & 34.4 & 34.4 & 38.3 & 39.4 \\

Abel-13B & 31.2 & 48.0 & 37.2 & 22.4 & 34.7 \\

Xwin-Math-13B & 45.9 & 29.8 & 33.1 & 30.2 & 34.8 \\

InternLM2-Math-Plus-20B & 38.3 & 52.9 & 37.8 & 18.4 & 36.9 \\

GPT-4o & 38.1 & 48.3 & 37.7 & 59.0 & 45.8 \\

Claude 3.7-Sonnet & 60.7 & 45.8 & 56.8 & 64.8 & 57.0 \\

Gemini 2.5-Pro & 60.4 & 48.6 & 56.7 & 77.0 & 60.7 \\

\midrule

Llama-3 + AMR (Ours) & 43.5 & 37.8 & 21.8 & 23.1 & 31.6 \\

InternLM2 + AMR (Ours) & 47.6 & 49.1 & 36.5 & 38.4 & 42.9 \\

Qwen2.5 + AMR (Ours) & 52.7 & 41.2 & 46.8 & 46.5 & 46.8 \\

\bottomrule\end{tabular}\end{adjustbox}\end{table}

\subsection{Main Results}

\paragraph{Field Ability.}

As shown in Table 1, AMR improves the field abilities of Llama-3-8B-Instruct. Beyond improving its Level 1 and 2 average accuracy to 38.6\% and 27.9\%, it achieves 54.3\% in Algebra and 45.2\% in Analysis on Level 1 tasks, outperforming math-specific baselines like Deepseek-math-RL-7B and NuminaMath-7B-TIR. AMR's enhancement is not limited to general models; it consistently boosts the performance of math-specialized models as well. Integrating AMR with the InternLM2-Math-Plus-7B enhances its average accuracy to 46.2\% (+8.4\%) on Level 1 tasks and 42.7\% (+6.2\%) on Level 2 tasks. Notably, it outperforms 13B specialized models in some domains: it achieves 48.4\% in Analysis (Level 1) and 38.8\% in Topology (Level 2), successfully surpassing Abel-13B (40.4\%) and Xwin-Math-13B (38.3\%), respectively. For Qwen2.5 empowered by AMR, the average accuracy increases to 68.1\% (+3.7\%) on Level 1 and 63.7\% (+4.4\%) on Level 2. This performance indicates that it outperforms across all open-source models; for instance, in Algebra Level 2, its score of 71.4\% far surpasses InternLM2-Math-Plus-20B (49.6\%) and Xwin-Math-13B (51.6\%). Impressively, even when compared with commercial models, it still remains remarkably competitive and even surpasses them in specific domains. For instance, its Analysis (Level 2) score of 73.5\% surprisingly exceeds all evaluated commercial models. Furthermore, its Geometry (Level 2) score of 53.8\% significantly outperforms both Claude 3.7 Sonnet (39.0\%) and GPT-4o (36.3\%).

\paragraph{Logic Ability.}

As shown in Table 2, AMR effectively enhances the logical reasoning capabilities of Llama-3-8B-Instruct. It raises the overall average to 31.6\%, driven by steady improvements in Concept Attribute (43.5\%) and Forward Reasoning (21.8\%). The enhancements also consistently extend to math-specialized models. Equipping InternLM2-Math-Plus-7B with AMR elevates its average logic score to 42.9\% (+6.1\%). Notably, it achieves a remarkable surge in Forward Reasoning, jumping from 23.7\% to 36.5\%. This enhancement enables it to comprehensively surpass the logic performance of larger open-source models, including InternLM2-Math-Plus-20B (36.9\%) and Xwin-Math-13B (34.8\%). With AMR, the average performance of Qwen2.5 improves by 7.4\% to 46.8\%. It shows strong growth in complex reasoning tasks, achieving 46.8\% (+12.4\%) in Forward Reasoning and 46.5\% (+8.2\%) in Backward Reasoning. Impressively, even when compared to commercial models, it stays very competitive and wins in specific areas. Most notably, it achieves an overall average score of 46.8\% and a Forward Reasoning accuracy of 46.8\%, successfully outperforming GPT-4o which scores 45.8\% and 37.7\% in the respective metrics.

\begin{table}[htbp]\centering
\caption{Comparison on GSM8K and MATH. Baseline is Llama-3-8B-Instruct.}\label{tab:gsm8k}
\begin{tabular}{lrr}\toprule
Method & GSM8K & MATH \\\midrule
Baseline & 79.6 & 34.7 \\
DPO & 80.4 & 35.2 \\
Step-DPO & 81.3 & 36.0 \\
Math-Shepherd & 81.0 & 35.8 \\
RSD & 81.5 & 36.5 \\
MathFimer & 82.0 & 37.0 \\
\midrule AMR (Ours) & 83.3 & 37.9 \\\bottomrule
\end{tabular}\end{table}

\paragraph{Results on Other Benchmarks.}

To further verify the effectiveness of AMR, we further evaluate AMR on GSM8K and MATH datasets comparing with other frameworks for enhancing mathematical reasoning in LLMs. As summarized in Table~\ref{tab:gsm8k}, AMR exhibits highly competitive performance, achieving absolute gains of 3.7\% on GSM8K and 3.2\% on MATH over the base model. These improvements clearly surpass 
those obtained by recent advanced methods such as MathFimer \citep{yan2026mathfimer} (+2.4\% and +2.3\%, respectively) and RSD \citep{liao2025rewardguided} (+1.9\% and +1.8\%, respectively).

\begin{table}[htbp]\centering
\caption{Ablation study on Qwen2.5-Math-7B-Instruct.}\label{tab:ablation}
\begin{tabular}{lrrr}\toprule
Method & Overall & Field Avg. & Logic Avg. \\\midrule
Base & 50.6 & 61.9 & 39.4 \\
Joint LoRA & 51.8 & 62.7 & 40.9 \\
Best Single LoRA & 53.4 & 63.8 & 43.0 \\
Static Composition & 52.7 & 63.3 & 42.2 \\
Dynamic Composition & 54.5 & 64.5 & 44.5 \\
AMR (Ours) & 56.4 & 65.9 & 46.8 \\\bottomrule
\end{tabular}\end{table}

\subsection{Ablation Study}

we conduct an ablation study on Qwen2.5-Math-7B-Instruct to evaluate their importance, as summarized in Table 4. We first evaluate Joint LoRA, which is trained on the mixture of all atomic datasets,improves the base model by 1.2\% overall, while the Best Single LoRA achieves a larger gain of 2.8\%, suggesting that atomic specialization is more effective than directly mixing heterogeneous mathematical abilities. We further compare different composition strategies. Static Composition learns a globally shared combination of atomic experts and already improves the Overall score from 50.6\% to 52.7\%, demonstrating that composing multiple atomic capabilities can produce effective reasoning synergy. Dynamic Composition further boosts performance by 1.8\% over Static Composition through support-dependent combination weights, indicating that different support sets benefit from different expert mixtures rather than a universal composition. Building on this observation, AMR introduces instance-level prototype routing to select a more suitable composition for each problem, achieving the best overall performance of 56.4\% in AtomicMath benchmark.

\begin{table}[htbp]\centering
\caption{Leave-One-Out (LOO) ablation of atomic expert modules.}\label{tab:loo}
\begin{tabular}{lrrr}\toprule
Model / Drop Module & Overall & Field Avg. & Logic Avg. \\\midrule
AMR (Ours) & 56.4 & 65.9 & 46.8 \\
\midrule
w/o Algebra & 55.7 & 64.8 & 46.6 \\
w/o Analysis & 55.4 & 64.3 & 46.5 \\
w/o Backward Rea. & 56.1 & 65.8 & 46.4 \\
w/o Concept & 52.7 & 63.0 & 42.4 \\
w/o Forward Rea. & 56.2 & 65.8 & 46.6 \\
w/o Geometry & 56.1 & 65.4 & 46.8 \\
w/o Topology & 56.3 & 65.8 & 46.8 \\\bottomrule
\end{tabular}\end{table}

\subsection{LOO Analysis of Atomic Experts}

To better understand the contribution of each atomic expert within AMR, we conduct a leave-one-out (LOO) ablation, where one expert is removed from the full expert pool at a time while keeping the rest of the framework unchanged. As shown in Table 5, the most critical component is Conceptual Understanding. Removing this expert causes the largest overall drop, from 56.4 to 52.7, and leads to a particularly severe degradation on Logic Avg., which falls from 46.8 to 42.4. This result indicates that conceptual reasoning plays a foundational role in coordinating the higher-level logical behaviors required by mathematical problem solving. Among the remaining experts, Analysis and Algebra also make substantial contributions. Excluding Analysis reduces overall performance to 55.4 and lowers Field Avg. from 65.9 to 64.3, suggesting that it is especially important for mathematical tasks. Removing Algebra similarly decreases performance to 55.7, confirming that symbolic manipulation remains a core ingredient in the composed reasoning process. By contrast, removing Forward Reasoning, Backward Reasoning, Geometry, or Topology leads to only minor changes, implying that these experts are helpful but more complementary in the presence of the full expert pool. Overall, the LOO analysis supports our hypothesis that atomic experts are not equally important: AMR derives most of its gains from a small subset of highly influential capabilities, especially concept and analysis.

\begin{table}[htbp]\centering
\caption{Hyperparameter sensitivity of AMR on Qwen2.5-Math-7B-Instruct.}\label{tab:hyper}
\begin{tabular}{lrrr}\toprule
Hyperparameter & Overall & Field Avg. & Logic Avg. \\\midrule
\multicolumn{4}{l}{\textit{Prototype Pool Size}} \\
Top-4 prototypes & 53.8 & 63.0 & 44.6 \\
Top-6 prototypes & 54.3 & 63.5 & 45.1 \\
All ($M=9$) prototypes & 56.4 & 65.9 & 46.8 \\
\midrule
\multicolumn{4}{l}{\textit{Margin Threshold ($\delta$)}} \\
$\delta=0$ & 55.9 & 65.4 & 46.4 \\
$\delta=0.02$ & 56.4 & 65.9 & 46.8 \\
$\delta=0.05$ & 56.1 & 65.7 & 46.5 \\
\midrule
\multicolumn{4}{l}{\textit{Training Set Size}} \\
256 samples & 55.4 & 65.1 & 45.7 \\
512 samples & 56.4 & 65.9 & 46.8 \\
2048 samples & 56.0 & 65.5 & 46.5 \\
All ($N=7439$) samples & 55.8 & 65.3 & 46.3 \\\bottomrule
\end{tabular}\end{table}

\subsection{Efficiency and Parameter Analysis}

\paragraph{Efficiency.} To evaluate inference efficiency, we test the latency on a 256-example subset of Qwen2.5-Math-7B-Instruct with a single NVIDIA GeForce RTX 3090 GPU. Compared with the baseline Dynamic Composition (budget=10) that costs 184.0 seconds for inference, our framework only consumes 66.7 seconds, achieving a remarkable 2.76$\times$ speedup. The significant efficiency improvement stems from abandoning online gradient-free search during inference. Instead of performing iterative optimization for each test instance, AMR pre-constructs a prototype pool in the offline training phase. At test time, it merely conducts one lightweight forward pass and Logistic Regression classifier to rapidly route the input question to the prototype configuration.

\paragraph{Hyperparameter Analysis.}

We conduct sensitivity experiments on three hyperparameters. For prototype pool size, we expand the pool from top-4 to all 9 prototypes yields improvement across all metrics, with overall accuracy rising from 53.8\% to 56.4\% and the Field Avg. from 63.0\% to 65.9\%, showing that higher prototype size could enhance the overall performance. On the pseudo-label margin threshold $\delta$, $\delta$ = 0 retains ambiguous labels that blur decision boundaries, while $\delta$ = 0.05 excessively filters valid instances, $\delta$ = 0.02 balances label quality and the number of retained samples, achieving best result. Regarding the size of training set, AMR attains 56.4\% overall accuracy with 512 samples, reducing to 256 samples leads to underfitting, whereas scaling up to the full set degrades accuracy to 55.8\%, indicating that excessive training data introduces noise with marginal overfitting. More details in the Appendix.

\section{Conclusion}

In this paper, we presented Atomic MathReasoner (AMR), a two-stage framework for enhancing mathematical reasoning in LLMs through instance-level composition of atomic mathematical capabilities. AMR first constructs a diverse set of compositional prototypes by performing gradient-free search over independently trained atomic LoRA experts, and then distills these prototypes into a lightweight router that efficiently selects a suitable expert composition for each input during inference. Extensive experiments on the AtomicMath benchmark, together with evaluations on GSM8K and MATH, demonstrate that AMR consistently improves the mathematical reasoning performance of both general-purpose and math-specialized LLMs. Comprehensive ablation studies further verify the effectiveness of atomic specialization, compositional prototype construction, and instance-level routing, highlighting the importance of adaptive expert composition for mathematical reasoning.


\bibliographystyle{plainnat}
\bibliography{references}

\end{document}